\documentclass[letterpaper]{article} 
\usepackage{aaai2027}  
\usepackage[hyphens]{url}  
\usepackage{graphicx} 
\usepackage{natbib}  
\usepackage{caption} 
\usepackage{algorithm}
\usepackage{algorithmic}

\usepackage{newfloat}
\usepackage{listings}
\DeclareCaptionStyle{ruled}{labelfont=normalfont,labelsep=colon,strut=off} 
\floatstyle{ruled}
\newfloat{listing}{tb}{lst}{}
\floatname{listing}{Listing}

\usepackage{booktabs}

\usepackage{booktabs}
\usepackage{amsmath}
\usepackage{amssymb}
\usepackage{array}
\usepackage{tabularx}

\newcolumntype{R}[1]{>{\raggedright\arraybackslash\hsize=#1\hsize}X}

\newcommand{\hd}[1]{\begin{tabular}[b]{@{}c@{}}#1\end{tabular}}

\newcommand{\yes}{\ensuremath{\bullet}}
\newcommand{\partmark}{\ensuremath{\triangle}}
\newcommand{\no}{\ensuremath{\circ}}

\title{Do Latent Channels Actually Communicate?\\A Causal Audit of Latent Multi-Agent LLM Communication}
\author{
    Huixiang Zhang\equalcontrib,
    Mahzabeen Emu\corresponding
}
\affiliations{
\begin{tabular}[t]{c@{\hspace{1.5cm}}c}
\textsuperscript{\equalcontrib}Georgia Institute of Technology &
\textsuperscript{\corresponding}Memorial University\\
North Ave & 100 Signal Hill Rd\\
Atlanta, GA 30332 USA & St.\ John's, NL A1C 5S7 Canada\\
hzhang985@gatech.edu & memu@mun.ca
\end{tabular}
}

\begin{document}

\maketitle

\begin{abstract}
Latent communication in large language model (LLM)-based multi-agent systems (MAS) transmits continuous internal representations instead of text, but greater representational capacity does not establish that the receiver uses task-relevant information. End-task performance alone also cannot reveal whether an observed effect depends on message presence, content generated for the evaluated example, or information supplied by a separate agent. We introduce a causal audit that applies controlled message replacements at the boundary where the sender-produced representation enters the receiver. Four message settings support five measurements of encoded sender information, receiver sensitivity to message presence and identity, the task value of example-specific content, and the additional value supplied by a separate agent. We apply the audit to latent relay with Qwen3-4B and Qwen3-8B on GSM8K, ARC-C, and MATH-500. On GSM8K, the Qwen3-4B overall performance effect of $-1.00$ percentage point decomposes into a $-6.17$-point effect retained by an other-example message and a $+5.17$-point effect attributable to example-specific content; both component directions reverse at 8B. On MATH-500, the Qwen3-4B gain of $15.00$ points comprises $8.33$ points retained by an other-example message and $6.67$ points attributable to example-specific content, while the 8B gain is dominated by the former component. Self-substitution comparisons further show that example-specific content and other-agent value are distinct. These results show that aggregate accuracy does not identify how a latent message affects the receiver and motivate controlled message comparisons as a standard evaluation for latent communication.
\end{abstract}


\section{Introduction}
Recent large language model (LLM)-based multi-agent systems (MAS) have begun to explore communication channels beyond natural language. Conventional MAS typically coordinate agents through text messages, but text communication requires an agent to project its internal computation into discrete token sequences. This discretization exposes only the sampled symbols while discarding alternative information represented during generation \cite{pham2023,du2026}. To overcome this limitation, recent studies have explored richer communication carriers, including probability-weighted embeddings, hidden states, and key-value (KV) caches \cite{pham2023,zou2026,du2026}.

We refer to these approaches collectively as latent communication: communication protocols that transmit continuous internal representations, such as embeddings, hidden states, or KV caches, without requiring intermediate decoding into natural language. These approaches share a common intuition: preserving more information at the communication boundary may provide the receiver with a richer signal than text exchange. They report improved task performance over text-based communication or single-agent baselines, often together with reduced communication cost or inference overhead \cite{pham2023,zou2026,du2026}.

A performance gain after communication can arise from multiple sources: sender-specific content transferred across the boundary, additional computation introduced by the communication process, context reuse, or redundant reasoning trajectories \cite{cemri2025}. Therefore, an end-task accuracy difference alone cannot identify whether latent communication achieves genuine information transfer. The distinction is analogous to causal analysis in neural networks, where observing a representation's correlation with an output does not establish that the representation mediates the computation \cite{vig2020,meng2022}. A latent channel may influence receiver behavior without the receiver using the sender-specific information contained in the message \cite{lowe2019}. The research question is whether transmitted information causally contributes to receiver behavior.

\begin{figure*}[t]
\centering
\includegraphics[width=\textwidth]{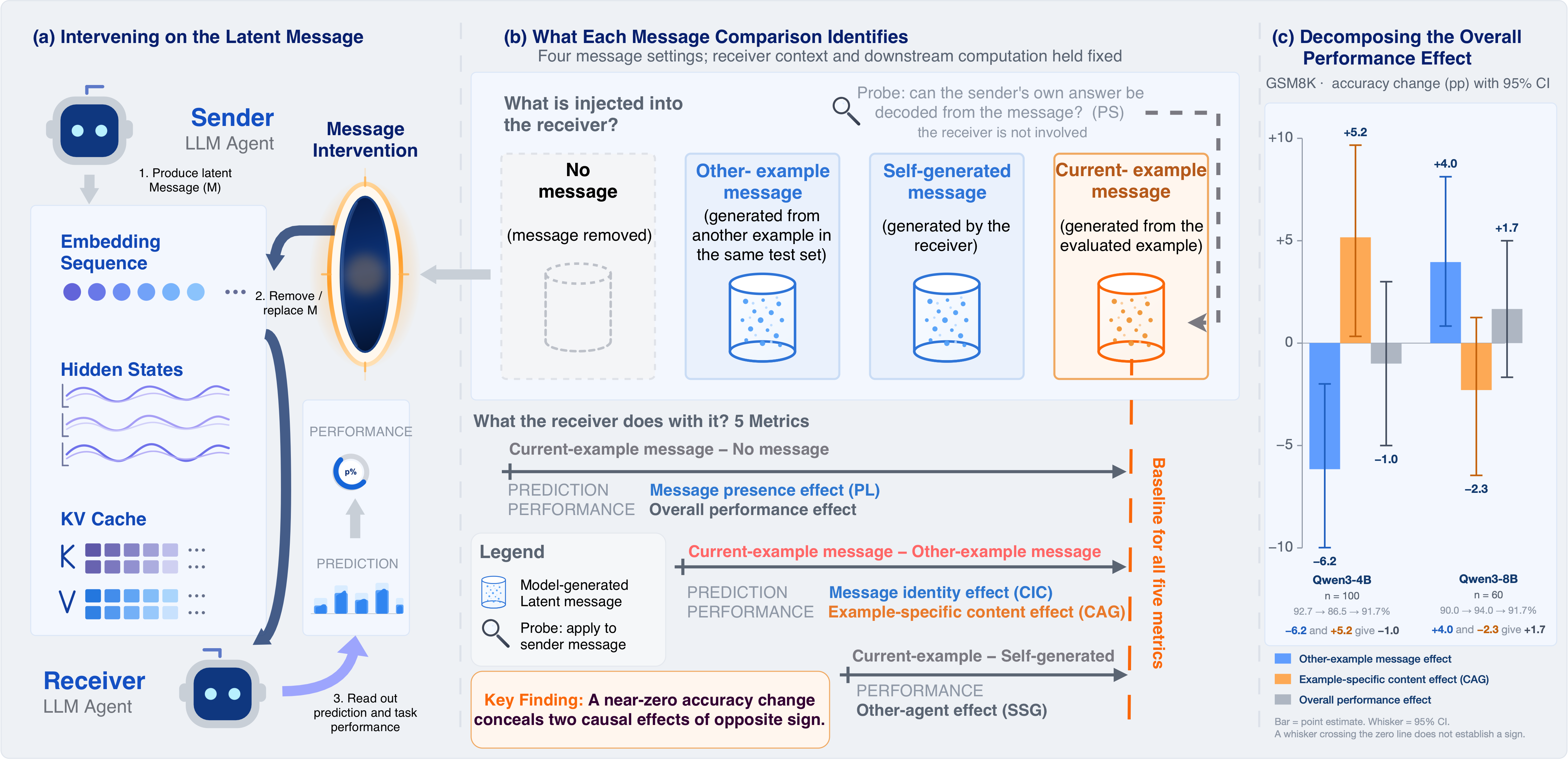}

\caption{A causal audit of latent communication.
(a) The sender produces a latent message $M$ as an embedding sequence, hidden states, or a KV cache; each intervention removes or replaces $M$ before it is injected into the receiver, while the receiver context and downstream computation are held fixed.
(b) An example is one test case from a benchmark test set. The four message settings are no message; an other-example message, generated by the sender through the same communication interface from a different example in the same test set and matched approximately in length; a compute-matched self-generated message; and the current-example message, generated from the evaluated example. Their comparisons define PL and CIC at the prediction-distribution level, and the overall performance effect, CAG and SSG at the task-performance level. PS instead tests whether the correctness of the sender's answer is decodable from $M$.
(c) On GSM8K with LatentMAS $\lambda_{H}$, the overall performance effect decomposes into the other-example message effect plus CAG: $-1.00 = -6.17 + 5.17$ percentage points (pp) for Qwen3-4B ($n = 100$) and $+1.67 = +3.96 - 2.29$ pp for Qwen3-8B ($n = 60$). Bars show point estimates and whiskers show 95\% confidence intervals. The near-zero 4B overall effect combines two opposing effects whose intervals exclude zero; both component estimates reverse sign at 8B; an interval crossing zero establishes neither a sign nor negligibility.}

\label{fig:overview}
\end{figure*}

Existing analyses of LLM-based MAS do not yet provide such an identification test. Failure taxonomies characterize recurrent coordination problems, while recent counterfactual studies intervene on individual agents, communication edges, or no-communication topologies to study error propagation and correlated agreement \cite{cemri2025,shen2025,huang2026,li2026}. These analyses establish whether communication helps or harms a system, but they do not determine which information inside a latent message causes the effect. Existing taxonomies classify latent-communication methods by the transmitted representation, sender--receiver alignment, and receiver-side fusion mechanism, but do not test whether the receiver uses content generated for the evaluated example \cite{liu2026}.

We introduce a causal audit that intervenes on the message before it enters the receiver while holding the receiver context and downstream computation fixed. We use example to mean one test case from a benchmark test set. The current-example message is the original message generated by the sender from the evaluated example. The first question is whether the message contains information about the sender. Positive signaling (PS) asks whether a declared sender variable is decodable from the message. In this work, the primary variable is whether the sender's own answer is correct. PS measures information encoded in the message, not whether the receiver uses that information. Encoded information establishes receiver use only when changing the message changes receiver behavior. Positive listening (PL) therefore compares the receiver's prediction distributions under the current-example message and no message. This comparison asks whether message presence affects the receiver at all, without yet identifying which part of the message produces the change.

To test whether content from the evaluated example matters, we replace the current-example message with an other-example message. An other-example message is generated by the sender from a different example in the same test set and is approximately matched in length. Because it is model-generated through the same communication interface, it preserves the original message structure, but its content comes from another example. Causal influence of communication (CIC) measures how the receiver's prediction distribution changes between the current-example and other-example messages. Content-attributable gain (CAG) measures the corresponding signed difference in task performance. CIC tests whether message identity affects receiver predictions, whereas CAG isolates the task value of content generated for the evaluated example. Content generated for the evaluated example may be useful without requiring a separate agent. The self-substitution gap (SSG) therefore compares the sender's current-example message with a self-generated message produced by the receiver through the same communication interface under a matched computation budget. This comparison asks whether a separate agent contributes task value beyond what the receiver can generate for itself.

Figure~\ref{fig:overview} summarizes the audit. Panel~(a) places embedding sequences, hidden states, and KV caches behind a shared message boundary, so every intervention removes or replaces the message before receiver injection. Panel~(b) presents four message settings: no message, an other-example message, a self-generated message, and the current-example message. Their comparisons separate message presence, message identity, example-specific content, and other-agent value, while PS is measured directly on the message. Panel~(c) shows that the overall performance effect decomposes exactly into the other-example message effect and CAG. The results illustrate why the decomposition is necessary. On GSM8K, the near-zero overall performance effect for Qwen3-4B combines two opposing components, while both component directions reverse for Qwen3-8B. Across models and tasks, similar overall performance can therefore arise from different communication mechanisms. We make the following contributions:
\begin{itemize}
\item We formulate latent communication as a capacity-versus-usage problem and provide a common causal audit for probability-weighted embeddings, latent-thought hidden states, KV caches.

\item We introduce a five-part measurement suite comprising PS, PL, CIC, CAG, and SSG. The suite separates information encoded in a message, receiver sensitivity to message presence and identity, the task value of example-specific content, and the value contributed by a separate agent.

\item We develop a standardized intervention design centered on four model-generated message settings, complemented by a preservation-ordered diagnostic ladder, component restoration, message-realism measurements, receiver-instability checks, and paired inference with positive, negligible-effect, and inconclusive outcomes.

\item We provide evidence that latent-channel behavior is heterogeneous: faithful relay can cause degradation in one operating regime, while another regime exhibits both a substantial content-attributable effect and a comparably content-independent effect. This decomposition changes the interpretation of the same end-task accuracy gain.
\end{itemize}

\section{Related Work}
\label{sec:related}

This section positions our audit against four lines of work: latent-communication methods, measurement of emergent communication, causal interventions on neural representations, and system-level analyses of MAS communication. Table~\ref{tab:related} summarizes the identification properties that distinguish these lines.

\begin{table*}[t]
\centering
\small
\setlength{\tabcolsep}{4pt}
\begin{tabular}{@{}llccccc@{}}
\toprule
Work & Object / level &
\hd{Latent\\object} &
\hd{Message-\\level\\intervention} &
\hd{Example-\\specific\\content} &
\hd{Matched\\compute /\\self-subst.} &
\hd{Equivalence-\\based\\inference} \\
\midrule
\multicolumn{7}{@{}l}{\emph{Block A: latent-channel proposals}} \\
\addlinespace[2pt]
CIPHER \cite{pham2023}       & soft tokens             & \yes & \no       & \no & \no       & \no \\
LatentMAS \cite{zou2026}     & hidden states, KV rows  & \yes & \partmark & \no & \partmark & \no \\
OBF \cite{li2026a}            & compressed KV relay     & \yes & \partmark & \no & \no       & \no \\
DroidSpeak \cite{liu2026a}    & prompt-KV reuse         & \yes & \partmark & \no & \no       & \no \\
C2C \cite{fu2025}            & fused KV                & \yes & \no       & \no & \no       & \no \\
Interlat \cite{du2026}       & adapter hidden states   & \yes & \no       & \no & \no       & \no \\
ThoughtComm \cite{zheng2025} & shared latent workspace & \yes & \no       & \no & \no       & \no \\
\addlinespace[2pt]
\midrule
\multicolumn{7}{@{}l}{\emph{Block B: causal and diagnostic analyses of MAS communication}} \\
\addlinespace[2pt]
\citet{lowe2019}             & RL discrete symbols        & \no & \partmark & \no       & \no & \no \\
\citet{jaques2019}           & RL messages                & \no & \yes      & \no       & \no & \no \\
MAST \cite{cemri2025}        & text traces, observational & \no & \no       & \no       & \no & \no \\
CAPE/TCTE \cite{shen2025}    & agent text outputs         & \no & \yes      & \no       & \no & \no \\
CAGE-CAL \cite{huang2026}    & agent answer graph         & \no & \yes      & \partmark & \no & \no \\
\addlinespace[2pt]
\midrule
This paper & latent message at boundary & \yes & \yes & \yes & \yes & \yes \\
\bottomrule
\end{tabular}
\caption{Identification properties of related work. Columns indicate whether each
work studies a latent inter-agent message, intervenes on the message itself,
separates the effect of content generated for the evaluated example from the
effect retained under an other-example message, compares against a
compute-matched self-generated alternative (self-subst.), and supports
negligible-effect conclusions through equivalence-based inference.
(\yes) satisfied; (\partmark) partial; (\no) absent.}
\label{tab:related}
\end{table*}

Latent inter-agent communication builds on work in single-agent latent reasoning, where Coconut feeds a model its own last hidden state back as the next input instead of a sampled token \cite{hao2025}. Multi-agent methods replace text messages with different continuous representations. CIPHER transmits the expectation of vocabulary embeddings under the sender's token belief \cite{pham2023}. LatentMAS generates latent thoughts as last-layer states and relays the sender's layer-wise KV cache as working memory \cite{zou2026}. DroidSpeak reuses prompt KV across same-architecture models to cut serving latency \cite{liu2026a}. C2C learns a projector and fusion module that injects a sender's KV cache into a target model \cite{fu2025}. Interlat trains a receiver-side adapter over compressed hidden states \cite{du2026}, and ThoughtComm fuses inferred latent thoughts through a shared workspace \cite{zheng2025}. A recent framework organizes these designs by carrier and injection mechanism \cite{liu2026}. These methods are evaluated primarily through end-task performance and, in several cases, communication or inference efficiency. Compression results further challenge a simple bandwidth explanation: compressed KV relay can match or exceed full relay on several benchmarks, indicating that the size of the transmitted representation does not map monotonically to downstream utility \cite{li2026a}. Table~\ref{tab:related}, block~A, locates these proposals by their identification properties.

Emergent-communication research has long distinguished information encoded by a sender from information used by a receiver. Referential-game studies induced protocols between neural agents \cite{lazaridou2017}, and measurement pitfalls followed quickly. \citet{lowe2019} distinguish positive signaling, where the message depends on the sender's state, from positive listening, where the receiver's behavior depends on the message, and show that widely used metrics can certify the former while the latter fails entirely. \citet{jaques2019} make listening causal, scoring a message by the counterfactual shift it induces in the receiver's policy. We adapt these constructs as PS, PL, and CIC for continuous latent messages between LLM agents. PS introduces an additional estimation problem because the message is continuous and high-dimensional. Distribution-free mutual-information lower bounds are limited by sample size, while variational estimators trade bias against variance \cite{mcallester2020,poole2019,song2020}. We therefore interpret signaling estimates as finite-sample lower bounds rather than direct measurements of the channel's full information content. Section~3 adapts these measurements to continuous pretrained-model representations transmitted without a token interface.

Our intervention design also builds on causal analyses of internal model representations. Causal mediation analysis introduced indirect-effect estimands for neural components \cite{vig2020}. ROME operationalized causal tracing through clean, corrupted, and restored forward passes \cite{meng2022}. Later work cataloged the design choices that make patching results trustworthy, including corruption type, metric choice, and a preference for in-distribution replacement over Gaussian noising \cite{zhang2024}, and documented off-distribution hazards when patching real LLMs \cite{yeo2025}. We inherit the protocol and its cautions: restoration sweeps are primary, and we accompany each intervention with checks of message-distribution similarity and receiver instability. The difference is the locus and the target. Patching work localizes circuits within one forward pass; we patch across agents, at the pre-injection boundary where one model's tensor enters another model's computation. The mediator is the relayed message itself, and the quantity of interest is a channel-level verdict about transmitted content, not a component map.

A parallel line analyzes communication failures and dependencies at the system level. MAST derives a fourteen-mode failure taxonomy from more than sixteen hundred annotated traces and reports that measured gains over single agents are often minimal, with inter-agent misalignment as one of its three top-level categories \cite{cemri2025}. CAPE and TCTE intervene on an agent's text output with a do-operation and measure how errors and insights propagate through communication topologies \cite{shen2025}. CAGE-CAL contrasts a post-communication agent graph with a matched no-communication counterfactual and shows that communication can produce correlated consensus, which vote-share confidence mistakes for evidence \cite{huang2026}. 

Together, these studies show that end-task performance and agent agreement are insufficient to identify content-based communication. Yet they analyze or intervene on text outputs or agent-level structures rather than the latent message crossing the sender--receiver boundary. Prior work therefore leaves three methodological strands separate: latent-channel design, causal listening measures for discrete protocols, and system-level counterfactual analysis of text communication. The missing test is a latent-boundary audit that intervenes on the transmitted message itself and separates example-specific content from the other-example message effect. It must also compare the sender's message with a compute-matched self-generated alternative and support equivalence-based conclusions when content or other-agent value is practically negligible. Our audit provides this identification test.

\section{The Audit Framework}
\label{sec:framework}

This section defines a common intervention boundary for latent communication and the measurements it supports. We first formalize the message pathway, then introduce four message settings, five audit metrics, component attribution, and the associated inference checks.

\subsection{Setup and Audited Boundary}
\label{sec:framework:setup}

We use \emph{example} to mean one test case from a benchmark test set. In the formal notation, an example and its associated contexts form an episode

\begin{equation}
e=(q,a^\star,c_S,c_R)\sim\mathcal{D},
\label{eq:episode}
\end{equation}

where $q$ is the question, $a^\star$ is the gold answer, and $c_S$ and $c_R$ are the sender and receiver contexts. The sender produces a latent message $M=\phi(e)$, and the receiver combines it with its context through the injection map $\psi$. For a replacement message $\widetilde M$, the receiver output is

\begin{equation}
y_{\widetilde M}(e)=R\!\left(\psi(c_R,\widetilde M)\right).
\label{eq:intervention}
\end{equation}

A message intervention replaces $M$ with $\widetilde M$ while holding the receiver context, model weights, prompt template, decoding procedure, and downstream computation fixed. The declared message must be the only inter-agent information path that changes. The same boundary applies to probability-weighted embeddings, hidden states, and KV caches used by existing latent-communication methods \cite{pham2023,zou2026,du2026}: intervention always occurs after sender-side construction and before receiver-side injection.

Operationally, each method exposes its native message representation at this boundary. For CIPHER, we replace the probability-weighted embedding sequence before it enters the receiver input. For hidden-state methods, we replace the transmitted hidden-state sequence before receiver-side adaptation or re-encoding. For KV-cache relay, we replace only sender-produced rows in the receiver's initial cache, leaving receiver-generated states unchanged. Current-example and other-example messages are generated through the same native interface and are approximately length-matched. The comparison therefore changes example identity while preserving the communication carrier, message structure, and receiver-side computation.

Table~\ref{tab:metrics} previews the audit questions and their identifying comparisons.

\begin{table}[t]
\centering
\footnotesize
\setlength{\tabcolsep}{3pt}
\renewcommand{\arraystretch}{1.15}
\begin{tabularx}{\columnwidth}{@{}>{\raggedright\arraybackslash}p{0.85cm} R{0.95} R{1.05} >{\raggedright\arraybackslash}p{1.70cm}@{}}
\toprule
\textbf{Metric} & \textbf{Question} & \textbf{Comparison} & \textbf{Estimand} \\
\midrule
PS  & Does $M$ encode sender information?
    & Message $M$ and sender variable $X$
    & $I(M;X)$ \\
\addlinespace[2pt]
PL  & Does message presence change receiver predictions?
    & Current-example message vs.\ no message
    & $\bar D_{\mathrm{cur},0}$ \\
\addlinespace[2pt]
CIC & Does message identity change receiver predictions?
    & Current-example vs.\ other-example message
    & $\bar D_{\mathrm{cur},\mathrm{oth}}$ \\
\addlinespace[2pt]
CAG & Does example-specific content add task value?
    & Current-example vs.\ other-example message
    & $\bar U_{\mathrm{cur}}-\bar U_{\mathrm{oth}}$ \\
\addlinespace[2pt]
SSG & Does a separate agent add task value?
    & Current-example vs.\ self-generated message
    & $\bar U_{\mathrm{cur}}-\bar U_{\mathrm{self}}$ \\
\midrule
\multicolumn{4}{@{}l}{\emph{Derived task-performance contrasts (exact identities)}} \\
\addlinespace[3pt]
OPE & What changes when the message pathway is added?
    & Current-example message vs.\ no message
    & $\bar U_{\mathrm{cur}}-\bar U_{0}$ \\
\addlinespace[2pt]
OME & What changes without evaluated-example content?
    & Other-example message vs.\ no message
    & $\bar U_{\mathrm{oth}}-\bar U_{0}$ \\
\addlinespace[2pt]
BME & Does same-benchmark content add value?
    & Other-example vs.\ other-benchmark message
    & $\bar U_{\mathrm{oth}}-\bar U_{\mathrm{xbench}}$ \\
\addlinespace[2pt]
DSC & How much of CAG can the receiver generate itself?
    & Self-generated vs.\ other-example message
    & $\bar U_{\mathrm{self}}-\bar U_{\mathrm{oth}}$ \\
\bottomrule
\end{tabularx}
\caption{Audit metrics and their identifying comparisons. OPE, OME, BME, and DSC denote the overall performance, other-example message, benchmark-match, and derived self-generated contrasts; $\mathrm{OPE}=\mathrm{OME}+\mathrm{CAG}$ and $\mathrm{CAG}=\mathrm{DSC}+\mathrm{SSG}$.}
\label{tab:metrics}
\end{table}

\subsection{Message Settings and Audit Metrics}
\label{sec:framework:metrics}

The audit uses four message settings. The current-example message is generated by the sender from the evaluated example. The other-example message is generated by the same sender from a different, approximately length-matched example in the same test set; it preserves the original message structure, but its content comes from another example. The self-generated message is produced by the receiver through the same interface under matched computation. The fourth setting supplies no message:

\begin{equation}
\begin{alignedat}{2}
M_{0}&=\varnothing, &\qquad
M_{\mathrm{cur}}&=\phi_S(e),\\[2pt]
M_{\mathrm{oth}}&=\phi_S(e'), &\qquad
M_{\mathrm{self}}&=\phi_R^{(c^\star)}(e).
\end{alignedat}
\label{eq:settings}
\end{equation}

where $e'\neq e$ and $c^\star$ is the matched message-generation budget. A length-matched other-benchmark message is used only for BME.

PS asks whether a finite sender variable $X$, here primarily sender answer correctness, is decodable from $M_{\mathrm{cur}}$. We estimate $\mathrm{PS}_X=I(M_{\mathrm{cur}};X)$ with a cross-fitted lower bound and compare it with a permutation reference. PS establishes encoded information, not receiver use.

For setting $a$, let $P_a$ be the receiver prediction distribution and $Y_a$ its parsed answer. Define

\begin{equation}
\bar D_{a,b}=\mathbb{E}_e[D(P_a,P_b)],
\qquad
\bar U_a=\mathbb{E}_e[U(Y_a)].
\label{eq:readouts}
\end{equation}

We use Jensen--Shannon divergence at the first completion token and along a teacher-forced reference continuation for prediction-level readouts. Task-level readouts use exact-match accuracy or the declared task-specific score. The receiver-use metrics are

\begin{equation}
\begin{aligned}
\mathrm{PL}  &= \bar D_{\mathrm{cur},0},
&\qquad
\mathrm{CIC} &= \bar D_{\mathrm{cur},\mathrm{oth}},\\[2pt]
\mathrm{CAG} &= \bar U_{\mathrm{cur}}-\bar U_{\mathrm{oth}},
&\qquad
\mathrm{SSG} &= \bar U_{\mathrm{cur}}-\bar U_{\mathrm{self}}.
\end{aligned}
\label{eq:metrics}
\end{equation}

PL measures the message-presence effect and CIC the message-identity effect. CAG isolates the task value of example-specific content, whereas SSG measures the additional task value supplied by a separate agent. The task-level contrasts satisfy

\begin{equation}
\begin{aligned}
\bar U_{\mathrm{cur}}-\bar U_0
  &=(\bar U_{\mathrm{oth}}-\bar U_0)+\mathrm{CAG},\\[2pt]
\mathrm{CAG}
  &=(\bar U_{\mathrm{self}}-\bar U_{\mathrm{oth}})+\mathrm{SSG}.
\end{aligned}
\label{eq:decomp}
\end{equation}

\subsection{Component Attribution}
\label{sec:framework:attribution}

CIC shows whether message identity changes receiver predictions, but not which components cause the change. Starting from an other-example message, we restore selected current-example components $C$ while leaving the rest fixed. Their contribution is summarized by

\begin{equation}
\mathrm{NLD}(C)
=\frac{\mathrm{LD}_{\mathrm{rest}(C)}-\mathrm{LD}_{\mathrm{oth}}}
{\mathrm{LD}_{\mathrm{cur}}-\mathrm{LD}_{\mathrm{oth}}},
\label{eq:nld}
\end{equation}

where $\mathrm{LD}$ is the teacher-forced answer log-probability difference. NLD near $1$ indicates substantial restoration and NLD near $0$ little restoration; it is evaluated only when the denominator exceeds the numerical reference floor. Components follow the message structure, such as sequence windows or layer by region by key-or-value cells.

\subsection{Inference and Validity Checks}
\label{sec:framework:inference}

All task-performance contrasts are paired at the example level. Multiple other-example messages are averaged within each example before aggregation. We report paired bootstrap confidence intervals together with sign-flip or exact paired tests. PS uses label permutations, PL and CIC use the numerical reference floor from repeated identical runs, and signed task-level effects are compared with zero. Practical negligibility is assessed with an equivalence test under a declared margin.

Synthetic messages, including noise, scrambling, and shuffling, are used only as diagnostics. Each intervention is accompanied by message-distribution similarity and receiver-instability measurements, and outputs with no valid parsed answer remain in the analysis with task score zero. Three hard checks validate the implementation: masking the message must reduce identity effects to the numerical floor, replacing an other-example message with the current-example message must yield zero CIC and CAG, and full restoration must give $\mathrm{NLD}=1$.

\section{Experiments and Results}
\label{sec:results}

\subsection{Experimental Setup}

We evaluate Qwen3-4B and Qwen3-8B on GSM8K \cite{cobbe2021}, ARC-Challenge (ARC-C) \cite{clark2018}, and MATH-500 \cite{lightman2024}. GSM8K and ARC-C retain two benchmarks from the LatentMAS evaluation \cite{zou2026} and cover open-form arithmetic reasoning and multiple-choice science question answering, respectively. MATH-500 adds a lower-accuracy setting of competition-level mathematics, reducing the risk that ceiling performance obscures communication effects. The paired analyses contain 100 and 60 GSM8K examples, 80 and 40 ARC-C examples, and 60 and 40 MATH-500 examples for the 4B and 8B models, respectively. All runs use one NVIDIA A40 GPU with 48GB of memory, 16 Intel Xeon Gold 6338 CPU cores, and 128GB of RAM under Ubuntu 24.04, Python 3.12, CUDA 12.6, PyTorch 2.11.0, and Transformers 5.13.0. 

We evaluated latent-step counts $m\in\{0,2,5,10,20,40\}$ and use $m=40$ in the reported experiments. This value follows the original method's reported $40$--$80$ high-performance range and selects its lower-compute endpoint; it was not selected using performance on the evaluated examples. Across the repository, decoding was evaluated with greedy sampling and $T=0.6$, generation caps of 256, 2048, and 3072 tokens, thinking mode enabled and disabled, and the default and aligned KV-relay variants. The reported experiments use nucleus sampling with $T=0.6$ and top-$p=0.95$, a 2048-token generation cap, thinking mode enabled, and the default KV-relay implementation to match the original method's main configuration. The number of other-example messages was fixed at $K=4$ rather than selected through tuning. Unless otherwise specified, the sender denotes one or more upstream agents other than the final-answer agent, the receiver denotes the agent that produces the final answer, and the audited communication boundary is the point at which the sender-produced message is handed to the receiver.

\begin{figure}[H]
\centering
\includegraphics[width=\columnwidth]{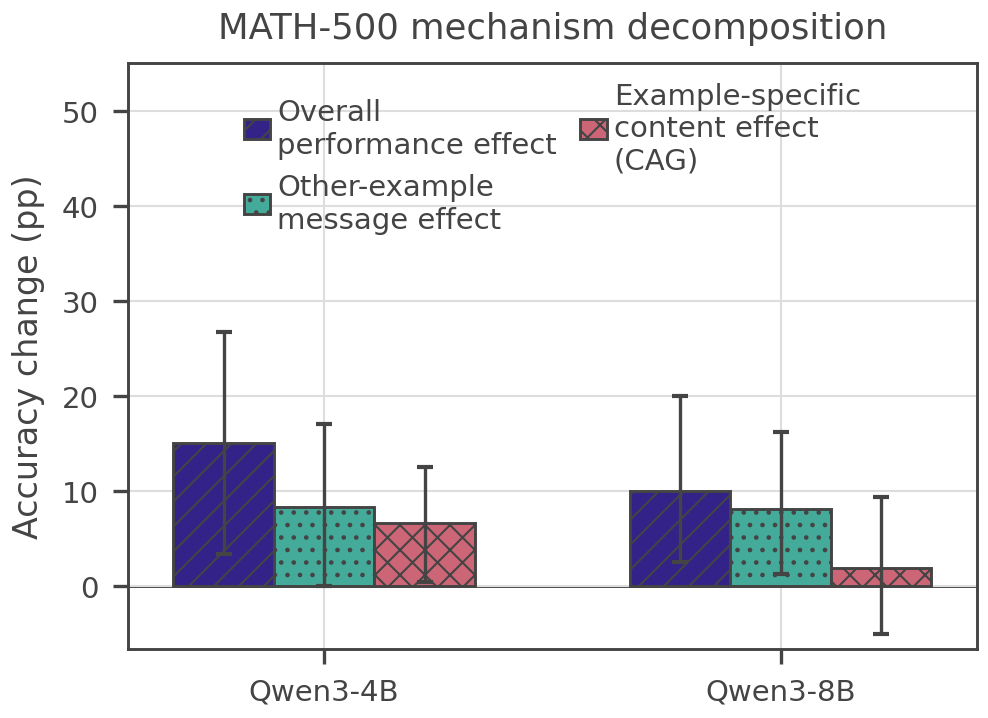}
\caption{MATH-500 decomposition of the overall performance effect into the other-example message effect and CAG. Bars show accuracy differences and whiskers show paired 95\% confidence intervals for Qwen3-4B ($n=60$) and Qwen3-8B ($n=40$).}
\label{fig:math_decomposition}
\end{figure}

Each model--benchmark configuration is evaluated on a fixed example set using independently seeded decoding runs. The Qwen3-4B and Qwen3-8B GSM8K results aggregate three and two seeds, respectively; the corresponding ARC-C results aggregate two and two seeds, and both MATH-500 configurations use one seed. These settings yield 300, 120, 160, 80, 60, and 40 example-draws for 4B GSM8K, 8B GSM8K, 4B ARC-C, 8B ARC-C, 4B MATH-500, and 8B MATH-500, respectively. Within each example-draw, the no-message, current-example, and self-generated settings are each executed once, while the other-example setting uses four independently assigned, length-matched messages. The four other-example outcomes are averaged within the evaluated example before paired aggregation. Runs across decode seeds are independent stochastic draws, whereas duplicate replays of an identical configuration reproduce the output bitwise. Accuracy differences are reported in percentage points (pp) with paired 95\% confidence intervals.

\subsection{Aggregate Performance Conceals Distinct Effects}

Figure~\ref{fig:overview}(c) shows the GSM8K decomposition. For Qwen3-4B, the overall performance effect is only $-1.00$ pp, but it combines a $-6.17$ pp other-example message effect with a $+5.17$ pp CAG. Both component intervals exclude zero. For Qwen3-8B, the overall effect remains small at $+1.67$ pp, while the component point estimates reverse direction: the other-example message effect is $+3.96$ pp and CAG is $-2.29$ pp. A near-zero aggregate effect can therefore conceal substantial and opposing communication effects.

MATH-500 produces a different decomposition. As shown in Figure~\ref{fig:math_decomposition}, the Qwen3-4B overall performance effect is $+15.00$ pp, comprising a $+8.33$ pp other-example message effect and a $+6.67$ pp CAG. The CAG confidence interval is $[0.42,12.50]$ pp. For Qwen3-8B, the $+10.00$ pp overall effect is dominated by the $+8.13$ pp other-example message effect, while CAG is $+1.88$ pp with an interval crossing zero. Thus, similar overall improvements can differ in how much they depend on content generated for the evaluated example.

ARC-C provides a lower-effect comparison. For Qwen3-4B, the $-0.63$ pp overall effect combines a $-2.97$ pp other-example message effect with a $+2.34$ pp CAG. The corresponding Qwen3-8B point estimates are $0.00$, $-1.56$, and $+1.56$ pp. Their intervals include zero, but the decomposition again separates effects hidden by the aggregate comparison.

\subsection{Example-Specific Content Does Not Imply Other-Agent Value}

\begin{figure}[t]
\centering
\includegraphics[width=\columnwidth]{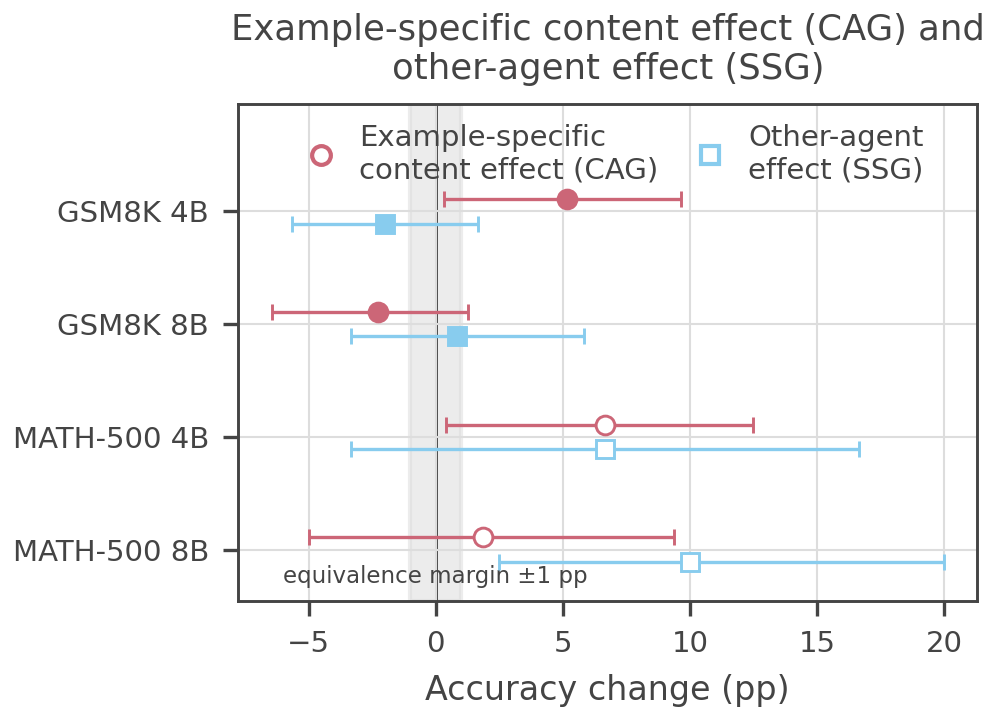}
\caption{CAG and SSG across GSM8K and MATH-500. Points show signed accuracy differences, whiskers show paired 95\% confidence intervals, and the shaded region marks the $\pm1$ pp equivalence margin.}
\label{fig:cag_ssg}
\end{figure}

CAG asks whether content from the evaluated example adds task value, whereas SSG asks whether that value requires a separate sender. Figure~\ref{fig:cag_ssg} shows that these questions can receive different answers. On GSM8K with Qwen3-4B, CAG is $+5.17$ pp, while SSG is $-2.00$ pp with an interval crossing zero. The sender's message therefore outperforms an other-example message, but not the receiver's compute-matched self-generated message. On MATH-500 with Qwen3-4B, CAG and SSG have the same point estimate of $+6.67$ pp, although the SSG interval crosses zero. For Qwen3-8B, CAG is only $+1.88$ pp, whereas SSG is $+10.00$ pp with a 95\% interval of $[2.50,20.00]$ pp. A separate agent can therefore add value even when the current-example versus other-example performance difference is small. None of the displayed intervals lies entirely within the $\pm1$ pp margin, so the data do not establish practical equivalence.

\subsection{Prediction Sensitivity and Task Value Are Distinct}

\begin{figure}[ht]
\centering
\includegraphics[width=\columnwidth]{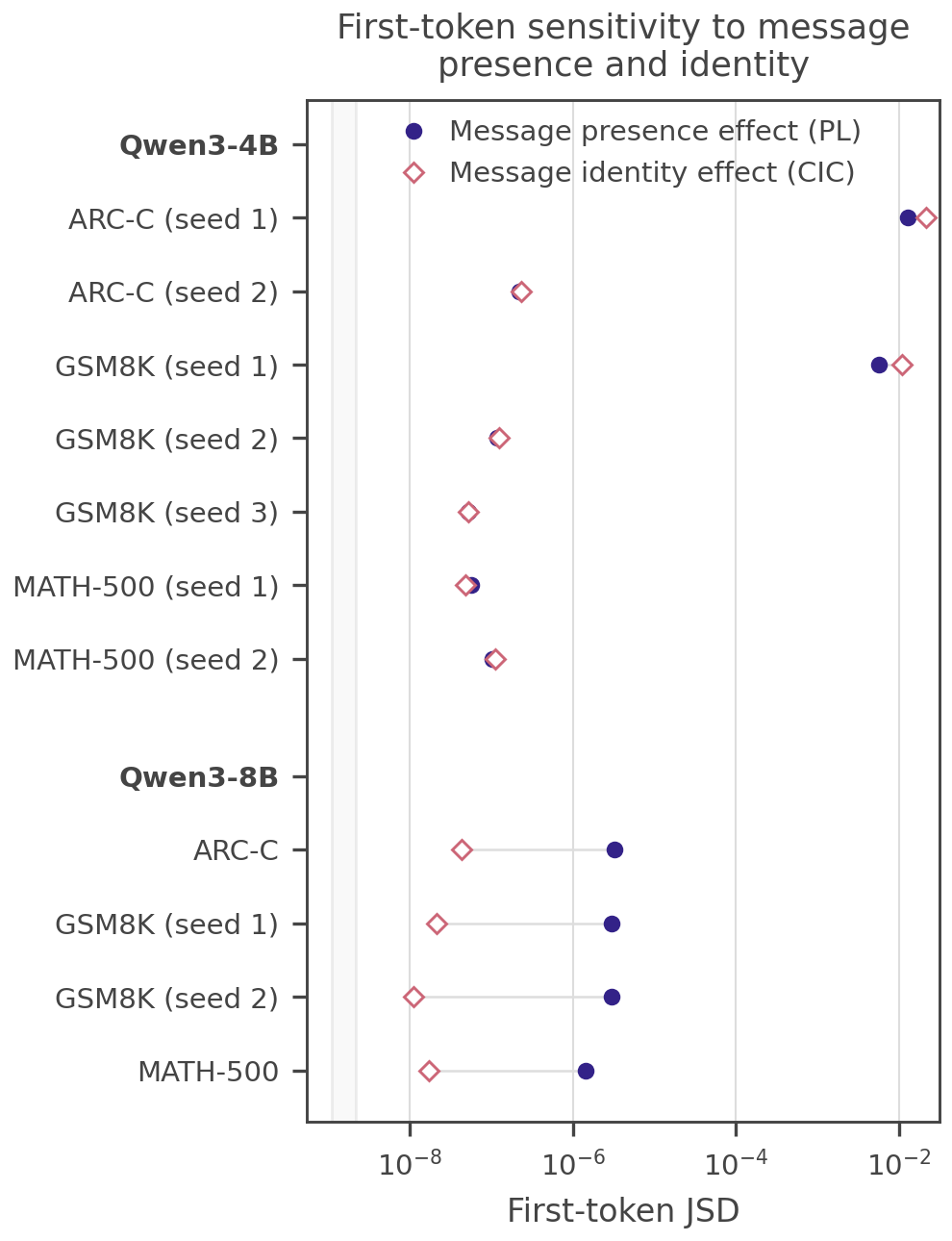}
\caption{First-token prediction sensitivity to message presence (PL) and message identity (CIC). Each pair reports Jensen--Shannon divergence for the same model, benchmark, and decode seed; the logarithmic axis emphasizes differences across scales rather than task-level utility.}
\label{fig:first_token}
\end{figure}

Figure~\ref{fig:first_token} compares first-token PL and CIC. The measured Jensen--Shannon divergences span several orders of magnitude. Several Qwen3-4B runs show large first-token responses to both message presence and message identity, whereas the plotted Qwen3-8B CIC values are consistently much smaller than their corresponding PL values.

This distributional pattern does not determine task performance. MATH-500 exhibits positive task-level effects even in runs with small first-token divergence, while large first-token sensitivity on GSM8K or ARC-C does not imply a positive overall performance effect. PL, CIC, CAG, and SSG provide complementary rather than interchangeable evidence.

\section{Conclusion}

We presented a causal audit for testing whether latent messages in LLM-based multi-agent systems carry information that the receiver actually uses. By intervening at a common sender--receiver boundary, the audit separates encoded sender information, sensitivity to message presence and identity, the task value of example-specific content, and the additional value supplied by a separate agent. Our results show that aggregate performance alone does not identify the mechanism of latent communication. Similar overall effects can arise from opposing components, while example-specific content and other-agent value can differ substantially across models and tasks. These findings motivate evaluating latent-communication methods through controlled message comparisons rather than a single end-task score.

\bibliography{aaai2027}


\end{document}